\pdfoutput=1

\documentclass[11pt]{article}

\usepackage{ACL2023}

\usepackage{times}
\usepackage{latexsym}
\usepackage{graphicx} 
\usepackage{booktabs} 
\usepackage{enumitem} 
\usepackage{float} 
\usepackage{array, makecell} %

\usepackage[T1]{fontenc}

\usepackage[utf8]{inputenc}

\usepackage{microtype}

\usepackage{inconsolata}

\title{UrduLLaMA 1.0: Dataset Curation, Preprocessing, and Evaluation in Low-Resource Settings}

\author{
  Layba Fiaz, Munief Hassan Tahir, Sana Shams, Sarmad Hussain \\
  Center for Language Engineering, \\
  Al-Khawarizmi Institute of Computer Science, \\
  University of Engineering and Technology, Lahore \\
  \texttt{\{firstname.secondname\}@kics.edu.pk}
}

\begin{document}
\maketitle
\begin{abstract}

Multilingual Large Language Models (LLMs) often provide suboptimal performance on low-resource languages like Urdu. This paper introduces UrduLLaMA 1.0, a model derived from the open-source Llama-3.1-8B-Instruct architecture and continually pre-trained on 128 million Urdu tokens, capturing the rich diversity of the language. To enhance instruction-following and translation capabilities, we leverage Low-Rank Adaptation (LoRA) to fine tune the model on 41,000 Urdu instructions and approximately 50,000 English-Urdu translation pairs. Evaluation across three machine translation datasets demonstrates significant performance improvements compared to state-of-the-art (SOTA) models, establishing a new benchmark for Urdu LLMs. These findings underscore the potential of targeted adaptation strategies with limited data and computational resources to address the unique challenges of low-resource languages.

\end{abstract}

\section{Introduction}
\label{sec:intro}

The field of language modeling has experienced a transformative shift, driven by the rapid evolution of Large Language Models (LLMs) that have set new standards in natural language understanding and generation. While proprietary models like OpenAI’s ChatGPT \citep{chatgpt} offer impressive capabilities, their closed nature restricts research accessibility. On the other hand, open models such as LLaMA\citep{llama3} and Mistral\citep{mistral7b} — though smaller in scale — have achieved competitive results across many languages. Nevertheless, both open and closed LLMs face significant challenges when applied to low-resource languages like Urdu. A primary hurdle is the inadequate representation of Urdu in training data, which results in limited vocabulary and poor encoding capabilities. This gap in data inclusion severely hampers the performance of LLMs on Urdu NLP tasks, as evidenced by recent benchmark studies\citep{34, 33}. Overcoming this limitation is crucial to unlocking the full potential of LLMs for Urdu and other underrepresented languages.
In this research, we tackle this challenge by developing an Urdu-specific LLM. We begin by continually pretraining Llama-3.1-8B-Instruct \citep{llama3} on 128 million Urdu tokens to enhance the model’s foundational representation of the language. This is followed by instruction fine tuning using 41,000 instructions to improve conversational capabilities, and additional fine tuning on 50,369 English–Urdu parallel sentence pairs to boost translation proficiency. The translation quality was evaluated using the BLEU score across three MT datasets, which showed that the UrduLLaMA 1.0 model outperformed the Llama-3.1-8B-Instruct base model. This trend was further validated through human evaluation, where two experts found that translations from the UrduLLaMA 1.0 model were more accurate than those generated by the base model.

The paper is structured as follows: Section \ref{sec:intro} introduces the study and Section \ref{sec:literature} presents the related work. Section \ref{sec:dataset} details the dataset curation and Section \ref{sec:filtering_pipeline} explains the steps taken to preprocess the data. This is followed by Section \ref{sec: UrduLLaMA 1.0}, which describes the development process of UrduLLaMA 1.0 including the experimental and training details, and Section \ref{sec:evaluation}, which covers the evaluation and discussion leading to Section \ref{sec:7}, which concludes the paper.

\section{Related Work}
\label{sec:literature}
Due to the lack of continual pretraining work on LLMs for Urdu, this section examines the closest related works. This includes models developed for Asian languages and low-resource languages built using the LLaMA framework.

Tamil-Llama \citep{tamilllama}, an Asian language model built on LLaMA 2 \citep{llama2}, incorporates 16,000 Tamil tokens and utilizes the Low-Rank Adaptation (LoRA) \citep{lora} technique for efficient training on Tamil datasets. The model was trained on an Nvidia A100 GPU with 80GB of VRAM for 48 hours, followed by instruct fine tuning on translated Alpaca datasets \citep{alpaca} and a custom subset of the OpenOrca \citep{openorca} dataset for 60 hours using Microsoft Azure’s Standard NC24 ads A100v4 instance. Performance evaluations indicate significant improvements in Tamil text generation, with the Tamil-Llama 13B model outperforming OpenAI's GPT-3.5-turbo on Tamil language tasks.

Taiwan-LLM \citep{18}, an LLM for Traditional Chinese, underwent continual pretraining on LLaMA 2 \citep{llama2} using 35.1 billion tokens and a diverse instruction set derived from 17 fine tuning datasets, including 20,000 user feedback instances. The training process leveraged the Transformer Reinforcement Learning (TRL) library \citep{15}, along with DeepSpeed ZeRO-2 \citep{16} and FlashAttention-2 \citep{17} to optimize memory usage and enhance training efficiency. Utilizing up to 48 NVIDIA H100 Tensor Core GPUs, Taiwan-LLM demonstrated superior performance in understanding and generating Traditional Chinese text, surpassing models such as GPT-4 and Claude-2.1.

PersianLLaMA \citep{persianllama}, the first large-scale Persian language model, was trained from scratch on 184 million tokens from Persian Wikipedia and 9 billion tokens from the OSCAR dataset \citep{19}. The training process leveraged DeepSpeed \citep{deepspeed} and TencentPretrain \citep{20}, two advanced frameworks for optimizing deep learning, utilizing two A100 GPUs with 80GB of VRAM over 12 days. Additionally, they conducted an experiment using LoRA \citep{lora} with the original English LLaMA weights, training on a single A100 GPU with 80GB of VRAM for over 70 hours. Their evaluations indicate that PersianLLaMA significantly outperformed its competitors in both understanding and generating Persian text. 

Airavata \citep{21} is an instruction-tuned model for Hindi, built by fine tuning OpenHathi \citep{22}, on 404k instruction instances from diverse Hindi instruction-tuning datasets. OpenHathi \citep{22} is again a model built on the LLaMA 2 7B architecture. The training employed both full fine tuning and supervised fine tuning using LoRA \citep{lora}. Their results demonstrated that Airavata significantly outperforms OpenHathi on most tasks, highlighting the effectiveness of fine tuning in aligning the base model to a variety of tasks. The details regarding their training infrastructure were not provided in their paper.

SeaLLMs \citep{seallm} is an innovative series of language models focused on Southeast Asian (SEA) languages. Built upon LLaMA 2 \citep{llama2} and Mistral 7B \citep{mistral7b}, SeaLLMs underwent continued pretraining with an extended vocabulary, followed by a hybrid approach for instruction and alignment tuning. Their evaluation claims that SeaLLMs significantly outperform ChatGPT-3.5 in non-Latin languages, such as Thai, Khmer, Lao, and Burmese, by large margins, while remaining lightweight and cost-effective to operate. The authors did not provide detailed information regarding the training infrastructure in their paper.

\begin{figure*}[!t]
    \centering
    \includegraphics[width=\textwidth]{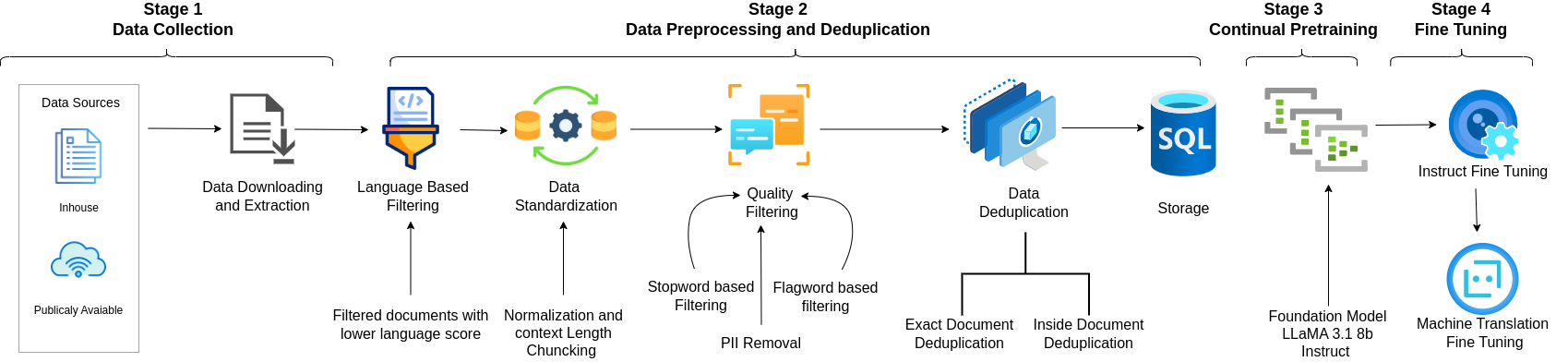} 
    \caption{Development of UrduLLaMA 1.0}
    \label{fig:example-image}
\end{figure*}

A study related to Chinese LLaMA \citep{24} extended different variants of LLaMA 2 \citep{llama2} by adding 20,000 Chinese tokens to the existing vocabulary. The model was pre-trained using LoRA \citep{lora} and fine tuned on Chinese instruction datasets formatted according to Alpaca \citep{alpaca}. Training was conducted on A40 GPUs (48GB VRAM), with up to 48 GPUs used depending on the model size. The parameter-efficient training with LoRA was carried out using the PEFT library\footnote{\url{https://github.com/huggingface/peft}}. Additionally, DeepSpeed \citep{deepspeed} was employed to optimize memory efficiency during training. Experimental results demonstrate that the newly proposed model significantly improves the original LLaMA's ability to understand and generate Chinese content.

Another study \citep{25} conducted a two-stage continual pretraining of LLaMA 3 8B \citep{llama3} for Chinese. Initially, they performed experiments on TinyLLaMA \citep{tinyllama}, and then applied their findings to train LLaMA 3 using 100B tokens, followed by fine tuning on Synthetic Scientific QA Data. The experiments were implemented using Hugging Face Transformers \citep{26}, incorporating Flash Attention and DeepSpeed ZeRO \citep{deepspeed} to optimize training efficiency. The study leveraged computing resources provided by the Public Computing Cloud at Renmin University of China. Their extensive experiments on a number
of evaluation benchmarks show that their approach can largely improve the performance of the backbone models, including both the general abilities and the scientific reasoning abilities without hurting the original capacities.

Latxa \citep{27} is a family of large language models for Basque ranging from 7 to 70 billion parameters. Latxa is based on Llama 2 \citep{llama2}, which they continue pretraining on their own Basque corpus comprising of 4.2B tokens.  The training of Latxa has been conducted using the GPT-Neox \citep{28} library. As infrastructure, they have leveraged the CINECA HPC Leonardo computing cluster located in Italy, which is powered by 3,456 nodes each containing 4x custom A100 64GB GPUs. They claimed that Latxa outperforms all previous open models they compared to by a large margin. 

\begin{table*}[ht]
    \centering
    \caption{Summary of Token Count Reduction Across Different Data Sources for Urdu-LLaMA}
    \label{tab:token_reduction}
    \renewcommand{\arraystretch}{1.5} 
    \resizebox{\textwidth}{!}{ 
        \begin{tabular}{lcccc}
            \hline
            \textbf{Source} & \textbf{Original Token Count} & \textbf{Token Count After Processing} & \textbf{Reduction} & \textbf{Percentage Reduction (\%)} \\
            \hline
            Publically Available Resources & 798,260,573 & 541,151,638 & 257,108,935 & 32.20 \\
            Inhouse          & 639,786,525 & 606,053,446 & 33,733,079  & 5.30  \\
            \hline

        UrduLLaMA 1.0 Dataset   & 1,438,047,098 & 1,147,205,084 & 290,842,014 & \textbf{-} \\
            \textbf{UrduLLaMA 1.0 Dataset  (in Billion)} & \textbf{1.43} & \textbf{1.14} & \textbf{0.29} & \textbf{-} \\
            \hline
        \end{tabular}
    }
\end{table*}
VinaLLaMA \citep{31}, an open-weight, state-of-the-art (SOTA) Large Language Model for the Vietnamese language, was built upon LLaMA 2 \citep{llama2} with an additional 800 billion trained tokens followed by fine tuning on 1 million sample instruction of Vietnamese and English. For our pretraining phase, they utilized a cluster consisting of eight nodes, each equipped with 8x Intel Habana Gaudi2 Processors. This phase was completed over the course of one week. In contrast, the fine tuning phase was conducted more rapidly, utilizing a single node of Google Cloud TPU v5e, and completed within a single day. They claim to achieve state-of-the-art results on different key benchmarks, showcasing fluency in Vietnamese and a deep understanding of their culture.

VI-MISTRAL-X \citep{29} is another LLM designed expressly for the Vietnamese language. It performed continual pretraining on 8 billion tokens selected from CulturaX \citep{7}, on the Mistral architecture, using 8 Nvidia H100 80GB SXM5
GPUs. Following the pretraining phase, vi-mistral-x underwent a series of fine tuning processes aimed at aligning the model
with specific NLP tasks. Through comprehensive testing
on various benchmarks, they have shown that vi-mistral-x has  outperformd existing Vietnamese LLMs in several
key areas.

This literature review presents the absence of dedicated Urdu LLMs highlights a significant need for such resources. This paper introduces the first continual pretraining followed by fine tuning of Llama-3.1-8B-Instruct model for Urdu, leveraging a dataset of 128 million tokens and using the LoRA \citep{lora} training approach for instruct tuning. This pioneering effort aims to pave the way for future research and development of more sophisticated Urdu language models.

\section{Dataset Curation}
\label{sec:dataset}

A pivitol challenge for building LLMs, particularly in low resource languages, is the availability of sizeable high-quality data for building foundation LLMs.  As the quality and diversity of data significantly influence the capabilities of LLMs \citep{1}, we supplemented our in-house dataset with data from several publicly available sources, including CC-100 \citep{5}, the Urdu corpus from OSCAR \citep{19}, the Urdu Web Corpus \citep{amir}, Urdu data from XLSum \citep{xlsum}, and \citep{words}. The raw text underwent a comprehensive pre-processing pipeline, outlined in Section \ref{sec:filtering_pipeline}, to ensure language-specific content, maintain quality, and remove duplicates. A summary of the collected datasets and the impact of processing is provided in Table \ref{tab:token_reduction}.

\section{Preprocessing Pipeline}
\label{sec:filtering_pipeline}
This section outlines the pre-processing steps applied to construct our dataset. We primarily adopted approaches similar to those proposed by \citep{2}, \citep{3}, and \citep{4}. The steps are as follows:

\begin{itemize}[leftmargin=5pt]

    \item \textbf{Language Filtering:}
This step was performed at the document level to retain only language-rich documents. Similar approaches have been adopted by others, such as the Falcon team \citep{6} for creating RefinedWeb, where they used the fastText language classifier from CCNet \citep{5} at the document level. This method has also been utilized by \citep{7} for building CulturaX and by \citep{8} for constructing the BigScience ROOTS corpus. In our case, we conducted language filtering using the CLE Urdu Language Identification API\footnote{\url{https://tech.cle.org.pk/api_langid}}, applying a threshold of 0.9 to ensure the retention of predominantly Urdu documents. To validate our choice of the CLE Urdu Language Identification API, we conducted experiments comparing it with fastText, assessing the effectiveness of both in identifying Urdu content within documents containing varying proportions of Urdu and non-Urdu text. The results of which are summarized in Table \ref{tab:lang_identification}, demonstrating that the CLE Urdu Language Identification API provided scores more aligned with the expected composition of test data.

    \item \textbf{Data Standardization:}
Data standardization involves the normalization and transformation of text data to make it more manageable and comprehensible during the model training process \citep{4}. Since syntax of the Urdu language requires specialized techniques, we applied as described in \citep{9}. Major steps include Unicode-based filtering, replacing non-standard characters with their standard forms, and handling Urdu-specific features such as end symbols, poetic symbols, and quotation marks. Additionally, some documents had varying lengths, so we split the text to maintain an average context length of 512 tokens.

\begin{table*}[h] 
\centering 
\caption{Language Identification Experiments with CLE Urdu Language Identification API and fastText} 
\label{tab:lang_identification} 
\renewcommand{\arraystretch}{1.2} 

\begin{tabular}{lp{4cm}p{4cm}} 
\hline 
\textbf{File Composition} & \textbf{CLE Urdu Language Identification API} & \textbf{fastText Model Score} \\ 
\hline 
80\% Urdu, 20\% non-urdu & 0.827 & lang:  ur , prob: 0.991 \\ 
50\% Urdu, 50\% non-urdu & 0.503 & lang:  ur , prob: 0.847 \\ 
25\% Urdu, 75\% non-urdu & 0.257 & lang:  en , prob: 0.439 \\ 
100\% Urdu & 1.000 & lang:  ur , prob: 0.997 \\ 
100\% Urdu (with urdu numerials) & 1.000 & lang:  ur , prob: 0.994 \\ 
Only Numericals & 0.000 & lang:  ru , prob: 0.349 \\ 
\hline 
\end{tabular} 

\vspace*{0.2cm}
\hspace*{0.1cm}\textit{Note: The "lang" represents the detected language, "prob" represents the probability score. \newline \hspace*{0.1cm} The language codes are as follows: "ur" = Urdu, "en" = English, "ru" = Russian.}
\end{table*}

    \item \textbf{Quality Filtering:}
To enhance the dataset’s quality, motivated by the data processing pipeline from \citep{8} and \citep{7}, we utilized various dataset metrics to identify and filter outlying documents. Filtering was applied based on stopword ratios, flagged word ratios, and empty documents. The threshold values for the stopword ratio and flagged word ratio were set at 0.1 and 0.025, respectively, which align with the threshold values used in the BigScience ROOTS project \citep{8}.

In addition to filtering, we implemented Personally Identifiable Information (PII) removal to protect sensitive data. We employed rule-based approach leveraging regular expressions  regexes library to detect and remove sensitive information such as phone numbers, identification numbers, and email addresses. These measures ensured that the dataset was free from personally identifiable information, enhancing privacy and usability for model training.

    \item \textbf{Deduplication:}

Despite thorough data cleaning, the remaining, dataset still contain a substantial amount of repeated data due to various reasons, including information being reposted on the web, multiple references to the same articles and plagiarism. The duplicated data can thus cause
memorization and significantly hinder generalization for LLMs \citep{11}. Therefore  deduplication is required as it decreases memorization of training data \cite{10}. 
Initially, deduplication was performed within individual datasets, followed by an overall deduplication across all datasets to address potential similarities among different sources. We applied deduplication at two levels with Table \ref{tab:deduplication} summarizes the results of this process:

\begin{enumerate} [leftmargin=15pt]

    \item \textbf{Exact Document Deduplication:}
We applied the SimHash technique, as used in the creation of WuDaoCorpora \citep{12} corpus, Roots Corpus for BigScience's BLOOMZ model \citep{13} and \citep{8}, to perform deduplication. A hash was generated from the content of each document (ignoring spaces) to uniquely identify it. If a duplicate hash was found, the corresponding document was removed.

    \item \textbf{Inside Document Deduplication:}
    
The second step involved deduplicating individual lines within the documents. Following the approach outlined by \citep{8}, we performed a line-by-line comparison to identify and remove repeated content. Duplicate lines, regardless of their position within a document, were eliminated.

\end{enumerate}

\begin{table}[h]
\centering
\caption{Impact of Deduplication on Dataset}
\label{tab:deduplication}
\renewcommand{\arraystretch}{1.2} 
\setlength{\tabcolsep}{4pt} 
\begin{tabular}{@{}lp{2.5cm}@{}}
\hline
\textbf{Step}                   & \textbf{Token Count} \\
\hline
Original Dataset                & 1.43 Billion \\
After Processing                & 1.14 Billion\\
After Overall Deduplication     & 1.08 Billion\\
\hline
\end{tabular}
\end{table}

\end{itemize}

\section{UrduLLaMA 1.0}
\label{sec: UrduLLaMA 1.0}
Llama-3.1-8B-Instruct, as introduced in \citep{llama3} by Meta, is built upon an extensive pretraining corpus of 15 trillion tokens. We leverage this model architecture for continual pretraining due to its open-source availability and the inclusion of Urdu language data in its training, making it a suitable choice for our research. The complete process of UrduLLaMA 1.0 making is illustrated in Figure \ref{fig:example-image}, follows four key stages: data collection, data processing, continual pretraining, and fine tuning, each playing a vital role in enhancing the model’s linguistic understanding and task adaptability.

\subsection{Continual Pretraining}
\label{sec:3.2}
The UrduLLaMA 1.0 model is trained on the Causal Language Modeling (CLM) task, enabling it to predict and generate the next word in a sequence. This stage plays a crucial role in refining LLaMA’s proficiency in Urdu by allowing the model to grasp the language’s intricate syntactic structures, semantic nuances, and unique linguistic traits. Leveraging its autoregressive nature, CLM mirrors the human process of language comprehension and generation, which is inherently context-dependent. Consequently, by the end of this initial training phase, LLaMA acquires the ability to generate and interpret Urdu text with contextual relevance and linguistic accuracy. 
\subsubsection{Pretraining Dataset}

Due to hardware limitations, 128 million tokens were used for continual pretraining of Llama-3.1-8B-Instruct \citep{llama3} from the curated dataset as explained in Section \ref{sec:dataset}. 

\subsubsection{Pretraining Setup}

The foundational model of UrduLLaMA 1.0 is initialized with the original Llama-3.1-8B-Instruct weights and want through pretraining using the \textit{bf16} precision setting. Our pretraining strategy involved full fine tuning, where all model parameters, including embeddings, LM heads, and attention weights, were trained. The training utilized an AdamW optimizer with a learning rate of 2e-5, and gradient accumulation steps of 1. Memory optimization techniques such as activation checkpointing, activation offloading, and memory-efficient FSDP wrapping were applied to manage the large model size effectively. Pretraining was conducted by utilizing 3 Nvidia L40 48GB GPUs. The model was trained for 3 epochs on the dataset, with the training process spanning approximately 2 weeks. The detailed parameters are described in Table \ref{tab:cpt}.
\begin{table}[ht]
    \centering
    \caption{Continual Pretraining Hyperparameters}
    \label{tab:cpt}
    \renewcommand{\arraystretch}{1.2} 
    \resizebox{\columnwidth}{!}{%
    \begin{tabular}{@{}lp{3cm}@{}}
        \hline
        \textbf{Configuration}     & \textbf{Value}        \\
        \hline
        Base Model                 & LLaMA 3.1             \\
        Parameters                 & 8B                    \\
        Training Tokens            & 128 Million           \\
        Epochs                     & 3                     \\
        Batch Size                 & 1                     \\
        Initial Learning Rate      & 2e-4                  \\
        Dropout Rate               & 0.1                   \\
        Max Sequence Length        & 512                   \\
        Training Precision         & FP16                  \\
        \hline
    \end{tabular}%
    }
\end{table}

\subsection{Instruct Tuning}
\label{sec:3.3}

Language models pre-trained using the causal language modeling (CLM) objective often struggle to follow user instructions and sometimes generate irrelevant or unintended content \citep{tamilllama}. This limitation arises because the CLM objective is designed to predict the next token in a sequence rather than understand or respond to instructions effectively \cite{35}. To address this issue and align the model’s behavior with user intentions, we employed instruction fine tuning. 
This step refines the LLM’s capabilities, allowing it to interpret and execute task-specific instructions more effectively in natural language. Rather than the traditional approach of adapting to specific datasets, instruction fine tuning focuses on a wide array of tasks articulated through language, ensuring the LLM’s adaptability without task-specific alterations. 

\subsubsection{Instruct Tuning Dataset}

We instruct tuned our model using 41,000 instructions, sources from two main sources. The first source included 26,000 instances from \citep{alpaca_translated}, a cleaned and translated version of the Stanford Alpaca dataset \citep{36}. This refined dataset addresses key issues such as hallucination, responses, ambiguous instructions, and low-quality samples, ensuring higher-quality training data. The second source comprises 15,000 translated instances from the Dolly dataset \cite{dolly_ur}, an Urdu translation of the original Dolly dataset \cite{dolly}, covering a wide range of NLP tasks related instructions.
\subsubsection{Instruct Tuning Setup}

For the instruction fine tuning, we incorporated the LoRA method \citep{lora}, where we integrated LoRA adapters into the attention vectors and subsequently trained the embeddings, LM heads, and the newly incorporated LoRA parameters. For the training infrastructure, we utilized Nvidia A100 40GB GPU. The detailed hyperparameters used for instructional fine tuning are listed in Table \ref{tab:ft parameters}.

\begin{table}[ht]
\centering
\caption{Fine Tuning Hyperparameters}
\label{tab:ft parameters}
\renewcommand{\arraystretch}{1.2} 
\setlength{\tabcolsep}{20pt} 
\begin{tabular}{@{}lp{2cm}@{}}
\hline
\textbf{Configuration} & \textbf{Value} \\
\hline
Training Data & 41,000 \\
Epoches & 3 \\
Batch Size     & 1 \\
Initial Learning Rate   & 2e-4 \\
Dropout Rate   & 0.1 \\
Max Sequence Length               & 512 \\
LoRA Rank                & 64 \\
LoRA Alpha               & 128 \\
Training Precision                & FP16 \\
\hline
\end{tabular}
\end{table}

\subsection{Fine Tuning on Machine Translation Task}
fine tuning is performed to adapt the pre-trained model, which has already learned general patterns and representations to specific task. Instead of focusing solely on specific linguistic pairs, fine tuning on machine translation datasets equipped our UrduLLaMA 1.0 to understand and translate text in a more sophisticated way, addressing domain-specific challenges. This process ensured that the UrduLLaMA 1.0 can effectively adapt to varied translation tasks, such as handling language-specific syntax, idiomatic expressions, and cultural nuances. 

We fine tuned the model using an in-house Machine Translation MT Corpus. This dataset is collected from various online sources covering diverse domains such as Banking\footnote{\url{https://www.sbp.org.pk/index.html}}, Law\footnote{\url{https://lgcd.punjab.gov.pk/}}, Weather\footnote{\url{https://nwfc.pmd.gov.pk/new/weekly-outlook-en.php}}, Agriculture\footnote{\url{https://aari.punjab.gov.pk/}}, and Food\footnote{\url{https://shireenanwer.com/recipes/main-course/zafrani-koftay/}}. The dataset comprises 62,970 entries, with a train-test split of 50,376 and 12,594 respectively. The model is fine tuned using the training split using parameters described in Table \ref{tab:ft parameters}.

\section{Evaluation on Machine Translation Task }
\label{sec:evaluation}

\subsection{Automatic Evaluation}

For automatic evaluation, we tested the model on a total of 16,299 instances, comprising test splits from the inhouse MT corpus, TICO-19 \citep{38}, and the Tatoeba Challenge \citep{37}. The TICO-19 dataset focuses on COVID-19-related domains such as health and public awareness, while the Tatoeba Challenge spans diverse domains for general-purpose translations. Specifically, the evaluation included 12,594 instances from the in-house test set, 2,042 instances from TICO-19, and 1,663 instances from the Tatoeba Challenge. This consistent test split was designed to align with SOTA models for a fair and reliable comparison.

Translation quality was assessed using the BLEU score \footnote{\url{https://www.nltk.org/_modules/nltk/translate/bleu_score.html}}, consistent with the metrics employed in SOTA model evaluation, providing an objective and standardized measure of the model’s performance on Urdu translations. The SOTA models evaluated are Llama-3.1-8B-Instruct \citep{llama3}, the base model used for continual pretraining; seamless-m4t-v2-large \citep{40}, a unified multilingual and multimodal translation model; and opus-mt-en-ur \citep{39}, a lightweight machine translation system tailored for low-resource languages, including Urdu. The results, showcasing a detailed comparison of translation quality across all models, are presented in Table \ref{tab:model_comparison}.

\begin{table}[ht]
    \centering
    \caption{Performance Comparison of Different Models on Machine Translation Datasets}
    \label{tab:model_comparison}
    \renewcommand{\arraystretch}{1.2} 
    \resizebox{\columnwidth}{!}{%
    \begin{tabular}{@{}lrrr@{}}
        \toprule
        \textbf{Model}        & \textbf{In-house} & \textbf{TICO-19} & \textbf{Tatoeba Challenge} \\ 
        \midrule
        UrduLLaMA 1.0   & \textbf{28.01} & 13.12 & 15.16  \\ 
        Llama-3.1-8B-Instruct          & 10.87             & 10.04            & 12.49                      \\
        opus-mt-en-ur                & 3.27              & 5.65            & 12.10                      \\
        seamless-m4t-v2-large          & 17.44             & \textbf{19.22}   & \textbf{22.76}             \\
        
        \bottomrule
    \end{tabular}%
    }
\end{table}

\begin{figure}[h] 
    \centering
    \includegraphics[width=0.5\textwidth]{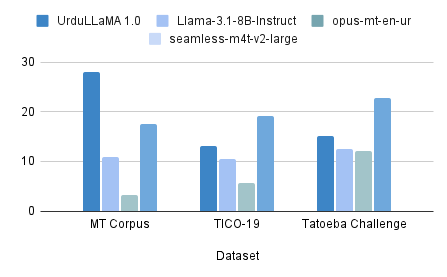} 
    \caption{Automatic Evaluation of Different Models on MT Datasets}
    \label{fig:chart_automatic}
\end{figure}

The results of this experiment are also confined in Figure \ref{fig:chart_automatic}. The results demonstrate that UrduLLaMA 1.0 significantly outperforms the base Llama-3.1-8B-Instruct model across all datasets, highlighting the effectiveness of our approach in improving translation performance for Urdu.
On the In-house dataset, UrduLLaMA 1.0 achieves the highest BLEU score, indicating its superior ability to handle domain-specific data. This suggests that fine tuning on Urdu text enhances the model’s ability to capture language-specific nuances and contextual meanings. In contrast, Llama-3.1-8B-Instruct shows lower performance, reinforcing that general-purpose multilingual models require adaptation for better Urdu translation. Interestingly, seamless-m4t-v2-large also performed well, though not surpassing UrduLLaMA 1.0, while opus-mt-en-ur struggles significantly, likely due to limited exposure to the dataset's domain.

For the TICO-19 dataset, UrduLLaMA 1.0 again surpassed Llama-3.1-8B-Instruct and opus-mt-en-ur, demonstrating that fine tuning allows the model to generalize better to unseen test sets. However, seamless-m4t-v2-large achieved the highest BLEU score, suggesting that massively multilingual pretrained models remain competitive on general purpose translation tasks and also indicating that pretraining alone is insufficient for high-quality Urdu translations.

In the Tatoeba Challenge dataset, UrduLLaMA 1.0 continues to outperform Llama-3.1-8B-Instruct and opus-mt-en-ur, demonstrating the impact of targeted fine tuning. However, seamless-m4t-v2-large achieves the best performance, likely due to its extensive multilingual training data. OpusMT shows improvement compared to the MT corpus dataset but remained behind UrduLLaMA 1.0 and seamless-m4t-v2-large, emphasizing the advantage of fine tuning over purely pretrained models.

These results highlight the importance of adapting general-purpose LLMs for specific languages. Fine tuning Llama-3.1-8B-Instruct on Urdu data has proven to be an effective approach, significantly improving translation quality over the base model. While seamless-m4t-v2-large remains a strong competitor due to its broad multilingual capabilities, UrduLLaMA 1.0’s strong performance on in-domain data suggests that domain-specific fine tuning is crucial for optimal results.

\subsection{Human Evaluation}

\begin{figure}[H] 
    \centering
    \includegraphics[width=0.5\textwidth]{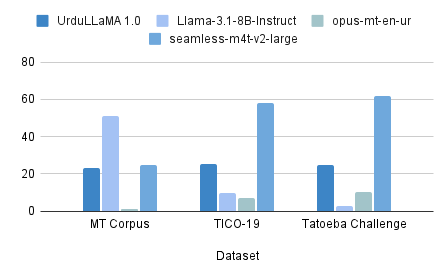} 
    \caption{Human Evaluation of Different Models on MT Datasets}
    \label{fig:chart_human}
\end{figure}

For the human evaluation of MT, we conducted a blind review. Two native Urdu linguists participated in this evaluation, where each was presented with an English source sentence along with translations from the four models, without any indication of model identity. They were instructed to select the translation they found most accurate and natural. A total of 300 test sentences were used with 100 sentences per dataset were randomly selected for manual evaluation. The preferences of both linguists were then aggregated to compare model performance. This evaluation provides valuable insight into the comparative quality of Urdu MT, reflecting the preferences of native Urdu experts in assessing translation accuracy and fluency summarized in Table \ref{tab:human} and interpreted in Figure \ref{fig:chart_human}. The difference in trends between human and automatic evaluation is due to the larger amount of testing data used in automatic evaluation. 

\begin{table}[ht]
    \centering
    \caption{Human Evaluation of Different Models on Machine Translation Datasets}
    \label{tab:human}
    \renewcommand{\arraystretch}{1.2} 
    \resizebox{\columnwidth}{!}{%
    \begin{tabular}{@{}lrrr@{}}
        \toprule
        \textbf{Model}        & \textbf{MT Corpus} & \textbf{TICO-19} & \textbf{Tatoeba Challenge} \\ 
        \midrule
        UrduLLaMA 1.0   & 23 & 25.5 & 24.5  \\ 
        Llama-3.1-8B-Instruct  & \textbf{51} & 9.5            & 3                     \\
        opus-mt-en-ur & 1  & 7            & 10.5                    \\
        seamless-m4t-v2-large & 25 & \textbf{58}   & \textbf{62}             \\
        
        \bottomrule
    \end{tabular}%
    }
\end{table}

The results show that seamless-m4t-v2-large provided better translations overall, and it is because of its specialized nature for translation and large amount of multilingual training data. However, 

UrduLLaMA 1.0, while a more general model, showed improvements over its base version in TICO-19 and Tatoeba Challenge. This progress is promising, as it suggests that with further refinement and more diverse training data, UrduLLaMA 1.0 has the potential to rival specialized models. The observed improvements in UrduLLaMA 1.0 are encouraging and highlight its adaptability and potential for high-quality translation in broader contexts.

\label{sec:6}

\section{Conclusion}
\label{sec:7}
In this paper, we introduced UrduLLaMA 1.0, a model specifically tailored for the Urdu language. We presented a comprehensive data processing pipeline to curate and prepare high-quality training data, addressing the challenge of limited publicly available Urdu datasets. UrduLLaMA 1.0 was continuoly pre-trained on a portion of this dataset using the Llama-3.1-8B-Instruct architecture, followed by instruction tuning to enable the model to understand and generate responses in a natural conversational format. This fine tuning leveraged a combination of the Alpaca and Dolly datasets. Furthermore, we performed fine tuning on a machine translation dataset to enhance the model's translation capabilities. Our evaluation results demonstrate that UrduLLaMA 1.0 outperforms its base model, exhibiting substantial improvements in machine translation tasks for Urdu. This work represents a significant step toward advancing the performance of LLMs for low-resource languages like Urdu and sets a new benchmark for future research in this domain.

\section*{Limitations}

Our model was trained on a limited portion of the Urdu dataset due to computational and cost constraints. As a result, it exhibits gaps in knowledge, particularly in capturing the nuances of Urdu culture and literature. While this version serves as a foundational step, its full potential can only be unlocked with access to a more extensive dataset to enhance its contextual understanding.

Additionally, detoxification processes were not incorporated during training, leaving the model uncensored and potentially prone to generating harmful or offensive content, which requires caution during deployment. 

Evaluating LLMs also presents a significant challenge, especially for underrepresented languages like Urdu, due to the lack of standardized benchmarks outside the European linguistic domain. Although this paper introduces a tailored evaluation approach for Urdu machine translation, it remains limited in scope and does not comprehensively assess the model's performance across diverse applications.
\section*{Ethics Statement}
This research utilizes publicly available, open-source datasets that do not contain personal or identifiable information, ensuring no associated risks. All work and ideas presented are original, with AI models used solely for grammatical correction and writing enhancement. Proper citations have been made for all models and datasets used.
Moreover as a generative model, it retains the potential to generate harmful or offensive content if prompted inappropriately, underscoring the need for responsible usage and careful oversight during deployment.
\bibliography{anthology,custom}
\bibliographystyle{acl_natbib}



\end{document}